%% file: 00main.tex
\documentclass[review]{elsarticle}

\usepackage[modulo]{lineno}
\modulolinenumbers[5]
\usepackage{comment}

\usepackage{comment}
\usepackage[colorlinks=true,urlcolor=blue,linkcolor=blue,citecolor=blue]{hyperref}
\usepackage{amsmath}
\usepackage{amssymb}
\usepackage{amsthm}
\usepackage{mathrsfs}
\usepackage{makecell}
\usepackage{multirow}
\usepackage{xcolor}
\usepackage{algorithm}
\usepackage{algpseudocode}

{
\theoremstyle{remark}

}

\makeatletter
\newenvironment{subtheorem}[1]{%
  \def\subtheoremcounter{#1}%
  \refstepcounter{#1}%
  \protected@edef\theparentnumber{\csname the#1\endcsname}%
  \setcounter{parentnumber}{\value{#1}}%
  \setcounter{#1}{0}%
  \expandafter\def\csname the#1\endcsname{\theparentnumber.\Alph{#1}}%
  \ignorespaces
}{%
  \setcounter{\subtheoremcounter}{\value{parentnumber}}%
  \ignorespacesafterend
}
\makeatother
\newcounter{parentnumber}
\newtheorem{theorem}{Theorem}

\newcommand{\mc}{\mathcal}
\newcommand{\mb}{\mathbb}
\newcommand{\mf}{\mathbf}
\newcommand{\mk}{\mathfrak}
\usepackage{bm}
\usepackage{upgreek}
\newcommand{\eps}{\varepsilon}

\usepackage{wrapfig}
\usepackage[justification=centering]{caption}
\usepackage{subcaption}
\usepackage{graphicx}
\usepackage{tabularx, booktabs}

\graphicspath{ {Plots/} }

\begin{document}

\begin{frontmatter}

\title{A Unified Hierarchical Multi-Task Multi-Fidelity Framework for Data-Efficient Surrogate Modeling in Manufacturing}

\author[uiuc]{Manan Mehta}
\author[umich]{Zhiqiao Dong}
\author[uiuc]{Yuhang Yang}
\author[umich,uiuc]{Chenhui Shao\corref{corrauth}}
\ead{chshao@umich.edu}
\ead[url]{https://me.engin.umich.edu/people/faculty/chenhui-shao/}

\address[uiuc]{Department of Mechanical Science and Engineering, 
         University of Illinois at Urbana-Champaign, Urbana, IL 61801, United States}
\address[umich]{Department of Mechanical Engineering, 
                University of Michigan, Ann Arbor, MI 41809, United States}
         
\cortext[corrauth]{Corresponding author}

\begin{abstract}
Surrogate modeling is an essential data-driven technique for quantifying relationships between input variables and system responses in manufacturing and engineering systems. Two major challenges limit its effectiveness: (1) large data requirements for learning complex nonlinear relationships, and (2) heterogeneous data collected from sources with varying fidelity levels. Multi-task learning (MTL) addresses the first challenge by enabling information sharing across related processes, while multi-fidelity modeling addresses the second by accounting for fidelity-dependent uncertainty. However, existing approaches typically address these challenges separately, and no unified framework simultaneously leverages inter-task similarity and fidelity-dependent data characteristics. This paper develops a novel hierarchical multi-task multi-fidelity (H-MT-MF) framework for Gaussian process-based surrogate modeling. The proposed framework decomposes each task’s response into a task-specific global trend and a residual local variability component that is jointly learned across tasks using a hierarchical Bayesian formulation. The framework accommodates an arbitrary number of tasks, design points, and fidelity levels while providing predictive uncertainty quantification. We demonstrate the effectiveness of the proposed method using a 1D synthetic example and a real-world engine surface shape prediction case study. Compared to (1) a state-of-the-art MTL model that does not account for fidelity information and (2) a stochastic kriging model that learns tasks independently, the proposed approach improves prediction accuracy by up to 19\% and 23\%, respectively. The H-MT-MF framework provides a general and extensible solution for surrogate modeling in manufacturing systems characterized by heterogeneous data sources.
\end{abstract}


\end{frontmatter}


\input{01introduction}

\input{02relatedwork}

\input{03modelformulation}

\input{04parameterestimation}

\input{05casestudies}

\input{06conclusion}

\section*{Declaration of Generative AI and AI-assisted technologies in the
writing process}

During the preparation of this manuscript, the authors used ChatGPT to help improve wording and correct English mistakes. After using this tool/service, the authors reviewed and edited the content as needed and take full responsibility for the content of the publication.

\input{appendix}

\bibliography{referencelist}

\end{document}

%% file: 01introduction.tex
\section{Introduction}

Surrogate modeling plays a critical role in manufacturing and engineering systems for quantifying relationships between process inputs and system responses, especially when first-principles physics models are unavailable, incomplete, or computationally expensive \cite{abio2024transfer,sun2026predicting}. By constructing empirical input-output mappings from experimental or simulation data, surrogate models enable process and design optimization \cite{kim2011process,bao2023surrogate,meng2020multi,yang2021hybrid}, uncertainty quantification \cite{bansal2022physics,li2019surrogate}, digital twin development \cite{waseem2025machine,chua2022surrogate,karkaria2024towards}, and fine-scale spatial and spatiotemporal interpolation \cite{Shao2017MTL, Mehta2021Adaptive, Haotian2021SpTmMTGP} in various industrial applications.

The effectiveness of surrogate modeling significantly depends on the quantity and quality of available data. Two major challenges arise when building surrogate models in manufacturing applications. First, learning complex and nonlinear input–output relationships often requires a large number of data points. Acquiring such data can be costly, time-consuming, or disruptive to production, particularly when experiments involve destructive testing, high-precision metrology, or high-fidelity simulations. Multi-task learning (MTL) offers a promising solution by enabling information sharing across multiple similar-but-not-identical tasks \cite{zhang2021survey}. By jointly learning correlated tasks, MTL reduces the data requirement for each individual task and improves learning performance in data-scarce scenarios.

Second, data in manufacturing and engineering systems are rarely homogeneous \cite{mehta2023greedy,meng2024meta,eslaminia2024federated}. Instead, they are often collected from heterogeneous sources with varying fidelity levels \cite{liu2025multi}. Examples include high-precision experiments combined with lower-cost sensing or modeling data \cite{bansal2023corrosion}, simulation models with coarse and fine discretizations \cite{zhang2020semiconductor}, reduced-order models coupled with detailed finite element simulations \cite{wang2020meta,baiges2020finite}, and metrology systems with different resolutions and precisions \cite{Suriano2015,shao2019engineering,yang2021data,dong2025fine}. These heterogeneous data sources differ not only in sampling density but also in uncertainty characteristics and noise levels. Multi-fidelity modeling techniques aim to account for such heterogeneity in order to improve prediction accuracy and robustness \cite{song2019radial,shi2020multi}.

While MTL and multi-fidelity modeling have independently advanced surrogate modeling, existing approaches typically address these two challenges separately. MTL models often assume homogeneous data quality across tasks, whereas multi-fidelity models usually focus on hierarchical relationships between fidelity levels within a single task. As a result, there is no unified framework that simultaneously leverages similarity across multiple related tasks and accounts for heterogeneous fidelity levels within and across those tasks.

To address this gap, we develop a novel hierarchical multi-task multi-fidelity (H-MT-MF) framework for Gaussian process (GP)-based surrogate modeling under heterogeneous data sources. The proposed approach decomposes each task’s response into two components: (1) a task-specific global trend and (2) a residual local variability component that is jointly learned across tasks using a hierarchical Bayesian formulation. Intrinsic uncertainty associated with different fidelity levels is incorporated through a heteroscedastic stochastic kriging (SK) formulation, which enables principled integration of heterogeneous data within and across tasks. The resulting framework accommodates an arbitrary number of tasks, design points, and fidelity levels while providing uncertainty quantification. The effectiveness of the proposed H-MT-MF framework is demonstrated using a simple 1D example and a real-world engine surface shape prediction case study. Compared to (1) MTL that does not account for fidelity information and (2) SK that learns tasks independently, the proposed approach improves prediction accuracy by up to 19\% and 23\%, respectively.

To the best of the authors' knowledge, the proposed H-MT-MF framework is the first unified formulation that simultaneously models cross-task similarity and fidelity-dependent intrinsic uncertainty. The conventional SK method is extended to a MTL setting, thereby enabling joint learning across multiple related processes while preserving rigorous predictive uncertainty quantification. In addition, we provide a rigorous probabilistic derivation of the H-MT-MF approach within a unified hierarchical Bayesian GP formulation, where both cross-task covariance/similarity and fidelity-dependent intrinsic variance are characterized in the joint likelihood. A customized expectation–maximization (EM) algorithm is further developed to enable efficient parameter estimation under this coupled structure. Finally, the effectiveness of the proposed framework is systematically evaluated through comprehensive case studies, demonstrating consistent performance improvements over state-of-the-art MTL and multi-fidelity modeling baselines.

The remainder of this paper is organized as follows. Section~\ref{sec:literature} reviews related work in multi-task and multi-fidelity Gaussian process modeling. Section~\ref{sec:modelformulation} presents the formulation of the proposed H-MT-MF framework. 
Section~\ref{sec:parameterestimation} describes the parameter estimation procedure. Section~\ref{sec:casestudies} demonstrates the proposed framework through a 1D illustrative example and a real-world manufacturing case study.
Finally, Section~\ref{sec:conclusion} concludes the paper and suggests future research directions.

%% file: 02relatedwork.tex
\section{Related Work} \label{sec:literature}

Surrogate modeling under limited and heterogeneous data has been studied in statistics, machine learning, and manufacturing communities. The proposed H-MT-MF framework lies at the intersection of three major research areas: MTL of GPs, multi-fidelity GP modeling, and heteroscedastic SK. We briefly review these areas and highlight the gap addressed in this work.

\subsection{Multi-Task Learning of Gaussian Processes}

MTL aims to improve learning performance by jointly learning multiple related tasks \cite{zhang2021survey}. In GP settings, multi-task models typically capture cross-task correlations through shared covariance structures or hierarchical Bayesian formulations \cite{Yu2005MTL, Bonilla2007MTL, Bonilla2009MTL}. Early work introduced kernel-based multi-task GP models that encode task relationships through structured covariance functions, e.g., \cite{Bonilla2007MTL}. Subsequent developments proposed hierarchical Bayesian approaches \cite{li2018hierarchical,Haotian2021SpTmMTGP} in which task-specific functions are modeled as samples from shared latent distributions, enabling information transfer across similar-but-not-identical tasks.

In manufacturing, MTL of GP models has been applied to a variety of applications including surface shape prediction in high-precision machining \cite{Shao2017MTL, Mehta2021Adaptive}, spatiotemporal modeling of tool condition progression \cite{Haotian2021SpTmMTGP}, solar panel performance forecasting \cite{Shireen2018Iterative}, response surface modeling for process optimizaiton \cite{yang2021hybrid}, manufacturing service allocation \cite{zhou2025collaborative}, corrosion prediction in multi-material systems \cite{bansal2025predicting}, and health monitoring of lithium-ion batteries \cite{guo2025robust}. These studies demonstrate that sharing information across correlated processes can significantly reduce data requirements and improve learning performance in data-scarce regimes.

Despite their effectiveness, existing multi-task GP models generally assume homogeneous data quality within each task. Observation noise is typically modeled as homoscedastic or task-specific but not explicitly linked to heterogeneous fidelity levels within tasks. As a result, these methods do not account for mixed-fidelity data arising from multiple sensing modalities, simulation resolutions, or metrology systems with varying precision.

\subsection{Multi-fidelity Gaussian Process Modeling}

Multi-fidelity modeling addresses scenarios where data are available at multiple levels of fidelity, such as low- and high-resolution simulations or measurements of varying precision. Multi-fidelity GP models provide a probabilistic framework for fusing such heterogeneous data while accounting for differences in uncertainty.

A classical and widely adopted approach to multi-fidelity GP modeling is autoregressive co-kriging \cite{kennedy2000predicting}, in which the response at a higher fidelity level is expressed as a scaled version of a lower fidelity response plus a discrepancy term modeled by an independent GP. This formulation enables hierarchical information transfer across fidelity levels and has been extensively used in simulation-based design and uncertainty quantification, e.g., \cite{zhou2020generalized,jia2025fast}. More general multi-fidelity GP models have been proposed to capture nonlinear relationships between fidelity levels, structured cross-correlations, and flexible discrepancy modeling. These approaches include nonlinear mapping strategies, deep Gaussian processes for multi-fidelity learning \cite{alvi2025accurate}, and Bayesian hierarchical models \cite{song2019general} that treat fidelity as an additional latent structure.

Multi-fidelity problems can also be viewed through the lens of heteroscedastic modeling \cite{le2005heteroscedastic}, where observations exhibit fidelity-dependent noise variance. In many practical settings, lower-fidelity data are characterized by higher uncertainty, while high-fidelity data provide more precise but often more expensive measurements. Heteroscedastic Gaussian process models account for input-dependent or source-dependent noise variance, thereby improving predictive robustness under heterogeneous uncertainty structures.

In simulation metamodeling, SK \cite{Ankenman2010StochasticKrig} extends classical kriging to incorporate intrinsic uncertainty arising from stochastic simulations. By modeling both extrinsic spatial correlation and intrinsic variance due to repeated simulations or measurements, SK enables heteroscedastic treatment of observation noise across design points. This formulation is particularly relevant in manufacturing contexts where measurement repeatability or simulation variance varies across operating conditions.

Despite these advances, existing multi-fidelity and heteroscedastic GP models are typically formulated for single-task settings. They focus on modeling relationships among fidelity levels within a single system and do not leverage cross-task similarity across multiple related processes.

\subsection{Summary of Research Gaps}

Existing methods address MTL and multi-fidelity or heteroscedastic modeling largely separately. Multi-task Gaussian process models enable information sharing across related processes but typically assume homogeneous data quality. Multi-fidelity and heteroscedastic GP models account for fidelity-dependent uncertainty but are generally restricted to single-task formulations. To the best of our knowledge, there is no unified framework that simultaneously (1) learns multiple correlated surrogate modeling tasks and (2) incorporates heterogeneous fidelity levels within and across tasks under a coherent hierarchical Bayesian formulation. The proposed H-MT-MF framework addresses this gap.

%% file: 03modelformulation.tex
\section{Model Formulation}\label{sec:modelformulation}

In this section, we first briefly review the SK model proposed in~\cite{Ankenman2010StochasticKrig} which characterizes both the intrinsic and extrinsic uncertainties in stochastic simulations. We then formulate the H-MT-MF model using a hierarchical Bayesian framework to learn multiple tasks simultaneously while accounting for intrinsic uncertainties.

\subsection{Review of Stochastic Kriging}\label{sec:stochkrig}

For deterministic experiments, Universal Kriging (UK) can learn an unknown response surface given a set of observations $\mf{Z} = \{ z_1,z_2,\dots,z_n \}^\top$ at design points $\mf{X} = \{\mf{x}_1,\mf{x}_2,\dots,\mf{x}_n \}^\top$, where $\mf{x}\in\mb{R}^d$. The model is represented as
\begin{equation}
    \label{eq:noiselessuniversalkriging}
    \mathsf{Z}(\mf{x}) = \mf{U}(\mf{x})^\top \bm{\upbeta} + \mathsf{M}(\mf{x}),
\end{equation}
where $\mf{U}(\mf{x})$ is a vector of known functions, $\bm{\upbeta}$ is a vector of unknown parameters of appropriate dimensions, and $\mathsf{M}$ is a zero-mean stationary Gaussian random field. $\mathsf{M}$ can be interpreted as being randomly sampled from a space of functions mapping $\mb{R}^d \rightarrow \mb{R}$. These functions exhibit a spatial correlation which is typically modeled by placing a GP prior on the functions, such that $\text{Cov}(\mathsf{M}(\mf{x}_i), \mathsf{M}(\mf{x}_k)) = \mk{R}(\mf{x}_i-\mf{x}_k; \bm{\uptheta})$, where $\mk{R}$ is the covariance that depends only on $\mf{x}_i - \mf{x}_k$. The hyperparameter vector $\bm{\uptheta}$ controls the smoothness of the underlying functions. This stochastic nature of $\mathsf{M}$ is referred to as extrinsic uncertainty, i.e., an uncertainty imposed on the model and not inherent to it. In reality, there is always intrinsic uncertainty associated with a stochastic model arising from the nature of the stochastic simulation (e.g., measurement noise). SK captures both uncertainties using the following model:
\begin{equation}
    \mc{Z}^j(\mf{x}) = \mf{U}(\mf{x})^\top \bm{\upbeta} + \mathsf{M}(\mf{x}) + \eps^j(\mf{x}) = \mathsf{Z}(\mf{x}) + \eps^j(\mf{x}),
    \label{eq:stochkrig}
\end{equation}
where $\eps^j(\mf{x})$ has zero mean and represents the intrinsic uncertainty. The superscript $j$ indicates the values obtained from the $j\textsuperscript{th}$ measurement replication at $\mf{x}$. The intrinsic errors $\eps^1(\mf{x}_i), \eps^2(\mf{x}_i), \dots$ at location $\mf{x}_i$ are i.i.d. $\mc{N}(0,\text{Var}[\eps(\mf{x}_i)])$, are independent of $\eps^j(\mf{x}_k) \hspace{0.5em}\forall k \neq i$, and are independent of $\mathsf{M}$. Here we note that SK allows for $\text{Var}[\eps(\mf{x})]$ to depend on location $\mf{x}$, i.e., the model allows for a heteroscedastic treatment of intrinsic variance across the design space.

Let $\overline{\bm{\mc{Z}}} = [\overline{\mc{Z}}(\mf{x}_1), \overline{\mc{Z}}(\mf{x}_2), \dots, \overline{\mc{Z}}(\mf{x}_n)]^\top$ denote the vector of sample means of the measurement responses, where
\begin{equation}
    \overline{\mc{Z}}(\mf{x}_i) = \frac{1}{n_i}\sum_{j=1}^{n_i}\mc{Z}^j(\mf{x}_i) = \mathsf{Z}(\mf{x}_i) + \frac{1}{n_i}\sum_{j=1}^{n_i}\eps^j(\mf{x}_i),
\end{equation}
and $n_i$ is the number of measurement replications at $\mf{x}_i$. Let $\bm{\Sigma}_\mathsf{M}$ be the $n\times n$ matrix of spatial covariances between the design points, where the $(i,k)\textsuperscript{th}$ entry $\bm{\Sigma}_\mathsf{M}(\mf{x}_i, \mf{x}_k)$ gives $\text{Cov}(\mathsf{M}(\mf{x}_i), \mathsf{M}(\mf{x}_k)) = \mk{R}(\mf{x}_i-\mf{x}_k; \bm{\uptheta})$. Consequently, $\bm{\Sigma}_\mathsf{M}(\mf{x}, \mf{x}_u)$ is the $n\times 1$ vector of covariances between an unobserved point $\mf{x}_u$ and the design points. Further, let $\bm{\Sigma}_\eps$ be the $n \times n$ matrix formed by the intrinsic noise such that the $(i,k)\textsuperscript{th}$ element of $\bm{\Sigma}_\eps$ is $\text{Cov}\left(\frac{1}{n_i}\sum_{j=1}^{n_i}\eps^j(\mf{x}_i),\frac{1}{n_k}\sum_{j=1}^{n_k}\eps^j(\mf{x}_k) \right)$. From the independence assumptions stated above, $\bm{\Sigma}_\eps$ reduces to a diagonal matrix $\text{Diag}\{\sigma_1^2/n_1,\sigma_n^2/n_2,\dots,\sigma_n^2/n_n\}$ where $\sigma_i^2 := \text{Var}(\eps^j(\mf{x}_i))$. The Mean Squared Error (MSE) optimal predictor for SK at the unobserved point $\mf{x}_u$ is given by
\begin{equation}
    \widehat{\mathsf{Z}}(\mf{x}_u) = \mf{U}(\mf{x}_u)^\top \bm{\upbeta} + \bm{\Sigma}_\mathsf{M}(\mf{x}, \mf{x}_u)^\top \big( \bm{\Sigma}_\mathsf{M} + \bm{\Sigma}_\eps \big)^{-1} \left( \overline{\bm{\mc{Z}}} - \mf{U}\bm{\upbeta} \right),
    \label{eq:mseoptimalpredictor}
\end{equation}
where $\mf{U} = \left( \mf{U}(\mf{x}_1)^\top,\mf{U}(\mf{x}_2)^\top,\dots,\mf{U}(\mf{x}_n)^\top \right)^\top$. An important result proven in~\cite{Ankenman2010StochasticKrig} is that by sensibly substituting $\widehat{\bm{\Sigma}}_\eps = \text{Diag}\{\widehat{\sigma}_1^2/n_1,\widehat{\sigma}_2^2/n_2,\dots,\widehat{\sigma}_n^2/n_n\}$ in place of $\bm{\Sigma}_\eps$, the MSE-optimal predictor in Eq.~\eqref{eq:mseoptimalpredictor} remains unbiased. Here $\widehat{\sigma}_i^2$ is the sample variance at $\mf{x}_i$, i.e., 
\begin{equation}
    \widehat{\sigma}_i^2 = \frac{1}{n_i-1}\sum_{j=1}^{n_i}\left( \mc{Z}^j(\mf{x}_i) - \overline{\mc{Z}}(\mf{x}_i) \right)^2.
\end{equation}
Prediction then follows by estimating of $\bm{\upbeta}$ and $\bm{\uptheta}$ using maximum log-likelihood.

\subsection{Multi-Task Multi-Fidelity Modeling Framework}\label{sec:mthgp}

To extend the SK model to learn $m$ response surfaces simultaneously, we rewrite Model~\eqref{eq:stochkrig} as
\begin{equation}
    \mc{Z}_l^j(\mf{x}) = \overbrace{\mf{U}_l(\mf{x})^\top \bm{\upbeta}_l}^{\text{global trend}} + \underbrace{\mathsf{M}_l(\mf{x}) + \eps_l^j(\mf{x})}_{\text{local variability}},
    \label{eq:stochkrigmtl}
\end{equation}
where the subscript $l$ indicates the task number, $l = 1,2,\dots,m$. The above model can be interpreted as a combination of a global trend and local spatial variability, where the local variability in turn comprises of both extrinsic and intrinsic components. The global trend is task-specific, whereas the local variability can be jointly learned across multiple similar-but-not-identical tasks. Like SK, we assume that the intrinsic errors $\eps^1_l(\mf{x}_i), \eps^2_l(\mf{x}_i), \dots$ at location $\mf{x}_i$ in task $l$ are i.i.d. $\mc{N}(0,\text{Var}[\eps_l(\mf{x}_i)])$, are independent of $\eps^j_h(\mf{x}_k) \hspace{0.5em}\forall k \neq i, h \neq l $, and are independent of $\mathsf{M}_l$. Through this assumption, the spatial variability across all tasks can be modeled together as a multi-task heteroscedastic zero-mean GP ${\eta}_l(\mf{x}) = \mathsf{M}_l(\mf{x}) + \eps_l(\mf{x})$.


We want to estimate $m$ correlated functions $\eta_l$, $l = 1,2,\dots,m$ on the training set $\{\mf{D}_l\}_{l=1}^m := \{\mf{X}_l, \bm{\upeta}_l\}_{l=1}^m$, where $\mf{X}_l \in \mb{R}^{n_l \times d}$ is the set of $d$-dimensional inputs for the $l\textsuperscript{th}$ task, $\bm{\upeta}_l \in \mb{R}^{n_l}$ are the values of $\eta_l$ on $\mf{X}_l$, and $n_l$ is the number of inputs for the $l\textsuperscript{th}$ task. Let $\mf{X}=\cup \mf{X}_l$. Note that there are a total of $n$ distinct data points in $\{\mf{D}_l\}_{l=1}^m$ with $\min(\{n_l\})\leq n \leq \sum_l n_l$. We first state two important results developed and proved in~\cite{Yu2005MTL} which form the basis of our  MTL model.

\begin{subtheorem}{theorem}
\begin{theorem}\label{theorem1}
    The mean $\bm{\upmu}$ and covariance $\bm{\Sigma}_\mathsf{M}$ for the GP prior on function values $\bm{\upeta} = [ \eta(\mf{x}_1),\dots,\eta(\mf{x}_n)]^\top$ are samples drawn from a Normal Inverse Wishart (NIW) distribution with scale matrix equal to a base kernel $\bm{\kappa} \succ 0$.
    \begin{equation}
        p(\bm{\upmu}, \bm{\Sigma}_\mathsf{M}) = \mc{N}\left(0,\frac{1}{\lambda}\bm{\Sigma}_\mathsf{M}\right)\mc{IW}(\nu, \bm{\kappa}).
        \label{eq:niw1}
    \end{equation}
\end{theorem}
\begin{theorem}\label{theorem2}
    Given $\bm{\upmu}$ and $\bm{\Sigma}_\mathsf{M}$ sampled from Eq.~\eqref{eq:niw1}, $\exists!$  $\bm{\upmu}_\alpha \in \mb{R}^n$, $\mf{C}_\alpha \in \mb{R}^{n \times n}$ such that $\bm{\mu} = \bm{\kappa}\bm{\mu}_\alpha$, $\bm{\Sigma}_\mathsf{M} = \bm{\kappa}\mf{C}_\alpha\bm{\kappa}$, and $\bm{\upeta} = \bm{\kappa\alpha}$ for $\bm{\alpha}\sim\mc{N}(\bm{\upmu}_\alpha,\mf{C}_\alpha)$, $\bm{\alpha} \in \mb{R}^n$.
\end{theorem}
\end{subtheorem}
It follows from Theorem~\ref{theorem2} that $\bm{\mu}_\alpha$ and $\mf{C}_\alpha$ also follow an NIW distribution, but with $\bm{\kappa}^{-1}$ as the scale matrix:
\begin{equation}
    \label{eq:niw2}
    p(\bm{\upmu}_\alpha, \mf{C}_\alpha) = \mc{N}\left(0,\frac{1}{\lambda}\mf{C}_\alpha \right)\mc{IW}(\nu, \bm{\kappa}^{-1}).
\end{equation}
Here, $\lambda$ and $\nu$ are the parameters specifying the NIW distribution. Using Eq.~\eqref{eq:niw2} as the hyperprior distribution of $\bm{\upmu}_\alpha$ and $\mf{C}_\alpha$, we describe the generative model in the following steps:
\begin{enumerate}
    \item $\bm{\upmu}_\alpha$ and $\mf{C}_\alpha$ are generated once using Eq.~\eqref{eq:niw2};
    \item For the true function $\eta_l$ underlying each task, we sample $\bm{\alpha}_l\sim\mc{N}(\bm{\upmu}_\alpha,\mf{C}_\alpha)$;
    \item For a design point $\mf{x}\in\mf{X}_l$, the response can be expressed as
    \begin{equation}
        \eta_l(\mf{x}) = \sum_{i=1}^n (\bm{\alpha}_l)_i \kappa(\mf{x}, \mf{x}_i) + \mc{V}_l(\mf{x}),
    \end{equation}
    where $\mf{x}_i\in\mf{X}$, $\mc{V}_l(\mf{x})\sim \mc{N}(0, \mathsf{V}_l(\mf{x}))$, and $\mathsf{V}_l(\mf{x}) \equiv \text{Var}(\eps_l(\mf{x})))$.
\end{enumerate}

The base kernel $\bm{\kappa}$ is a finite-dimensional realization of a base kernel function $\kappa(\cdot, \cdot)$ which describes the properties of the hyperprior. While the base kernel $\bm{\kappa}$ is solely restricted to be positive definite, we can choose a base kernel whose form resembles $\kappa(\mf{x}_i, \mf{x}_k) = \mk{R}(\mf{x}_i-\mf{x}_k; \bm{\uptheta})$. As seen from Theorem~\ref{theorem2}, the true GP kernel $\bm{\Sigma}_\mathsf{M}$ on set $\mf{X}$ can be recovered as $\bm{\kappa}\mf{C}_\alpha\bm{\kappa}$. Thus, the base kernel helps specify the extrinsic uncertainty across the design points in the MTL framework. To account for the intrinsic uncertainty, we can specify an $n_l \times n_l$ diagonal intrinsic noise matrix for task $l$ as $\bm{\Sigma}_{\eps,l} = \text{Diag}\{\mathsf{V}_l(\mf{x}_1)/n_{1,l}, \mathsf{V}_l(\mf{x}_2)/n_{2,l},\\ \dots, \mathsf{V}_l(\mf{x}_{n_l})/n_{n_l,l}\}$, where $n_{i,l}$ is the number of replications of the $i\textsuperscript{th}$ point in the $l\textsuperscript{th}$ task.

Combining the intrinsic and extrinsic uncertainties, we can derive a posterior variance measure at a predicted point $\mf{x}_u$ in task $l$, which helps quantify the prediction uncertainty at a test point. The prediction uncertainty for the noiseless UK model~\eqref{eq:noiselessuniversalkriging} is well studied. Stein~\cite{Stein1999Interpolation} showed that the MSE of the best linear unbiased predictor (BLUP) for Model~\eqref{eq:noiselessuniversalkriging} at predicted point $\mf{x}_u$ is given by
\begin{equation}
    \label{eq:noiselessuniversalkrigingvariance}
    \text{MSE}(\widehat{\mathsf{Z}}(\mf{x}_u)) = \overbrace{\bm{\Sigma}_\mathsf{M}(\mf{x}_u, \mf{x}_u) - \bm{\Sigma}_\mathsf{M}(\mf{x}, \mf{x}_u)^\top \bm{\Sigma}_\mathsf{M}^{-1} \bm{\Sigma}_\mathsf{M}(\mf{x}, \mf{x}_u)}^{\text{from residuals}} + \underbrace{\bm{\zeta}^\top \left( \mf{U}^\top \bm{\Sigma}_\mathsf{M}^{-1} \mf{U} \right)^{-1} \bm{\zeta}}_{\text{from trend}},
\end{equation}
where $\bm{\zeta} = \mf{U}(\mf{x}_u) - \mf{U}^\top \bm{\Sigma}_\mathsf{M}^{-1} \bm{\Sigma}_\mathsf{M}(\mf{x}, \mf{x}_u)$. Thus, the UK uncertainty incorporates both the prediction error variance of the residuals and the estimation error variance of the trend. Ankenmann et al.~\cite{Ankenman2010StochasticKrig} extended this to incorporate the heteroscedastic intrinsic variance across design points and formulated the MSE for Model~\eqref{eq:stochkrig} by replacing $\bm{\Sigma}_\mathsf{M}^{-1}$ in Eq.~\eqref{eq:noiselessuniversalkrigingvariance} with $(\bm{\Sigma}_\mathsf{M} + \widehat{\bm{\Sigma}}_\eps)^{-1}$, where $\widehat{\bm{\Sigma}}_\eps$ is estimated from the data. Since we learn the residuals using the hierarchical MTL formulation, the true covariance function $\Sigma_\mathsf{M}(\cdot, \cdot)$ is unknown. Mehta and Shao~\cite{Mehta2021Adaptive} showed that the variance measure can be approximated by weighing the learned kernel $\bm{\kappa}\mf{C}_\alpha \bm{\kappa}$ and base kernel $\kappa$ by their corresponding equivalent sample sizes as
\begin{equation}
    \Sigma_\mathsf{M}(\mf{x}_i, \mf{x}_j) \approx \frac{[m \bm{\kappa}(\cdot, \mf{x}_i)^\top \mf{C}_\alpha \bm{\kappa}(\cdot, \mf{x}_j) + \nu\kappa(\mf{x}_i, \mf{x}_j)]}{m + \nu},
\label{eq:compositekernel}
\end{equation}
where $ \bm{\kappa}(\cdot, \mf{x}_i) = \bm{\kappa}(\mf{X},\mf{x}_i) = [ \kappa(\mf{x}_1, \mf{x}_i),\dots,\kappa(\mf{x}_n, \mf{x}_i) ]^\top$. Using this empirical extrinsic covariance, we can write a variance measure at predicted point $\mf{x}_u$ in task $l$ as
\begin{equation}
    \label{eq:mtskvariance}
    \mb{V}(\widehat{\mathsf{Z}_l}(\mf{x}_u)) = \overbrace{\bm{\Sigma}_\mathsf{M}(\mf{x}_u, \mf{x}_u) - \bm{\Sigma}_\mathsf{M}(\mf{X}, \mf{x}_u)^\top \bm{\Sigma}^{-1} \bm{\Sigma}_\mathsf{M}(\mf{X}, \mf{x}_u)}^{\text{from residuals across all tasks}} + \underbrace{\bm{\zeta}_l^\top \left( \mf{U}_l^\top \bm{\Sigma}_l^{-1} \mf{U}_l \right)^{-1} \bm{\zeta}_l}_{\text{from task-specific trend}},
\end{equation}
where $\bm{\zeta}_l = \mf{U}_l(\mf{x}_u) - \mf{U}_l^\top \bm{\Sigma}_l^{-1} \bm{\Sigma}_\mathsf{M}(\mf{X}_l, \mf{x}_u)$, $\bm{\Sigma} = \bm{\Sigma}_\mathsf{M}(\mf{X}, \mf{X}) + \widehat{\bm{\Sigma}}_\eps$, and $\bm{\Sigma}_l = \bm{\Sigma}_\mathsf{M}(\mf{X}_l, \mf{X}_l) + \widehat{\bm{\Sigma}}_{\eps,l}$. Note that $\widehat{\bm{\Sigma}}_\eps = \bigoplus\limits_{l=1}^m \widehat{\bm{\Sigma}}_{\eps,l}$ is the intrinsic noise matrix for all design points across all tasks. Following the same logic as the UK MSE from Eq.~\eqref{eq:noiselessuniversalkrigingvariance}, the variance measure in Eq.~\eqref{eq:mtskvariance} is a combination of the prediction variance due to the MTL formulation and the variability due to estimating task-specific $\bm{\upbeta}_l$. Stated differently, the prediction variance at an unobserved point in a particular task is the combination of (1) variance of the residuals from all tasks which are learned together, and (2) variance of the estimation error of the task-specific learned trend.

Finally, assuming all the parameters for the model have been estimated (we discuss parameter estimation separately in Section~\ref{sec:parameterestimation}), the prediction at an unobserved point $\mf{x}_u$ in task $l$ is given by
\begin{equation}
    \widehat{\mathsf{Z}}_l(\mf{x}_u) = \mf{U}_l(\mf{x}_u)^\top \widehat{\bm{\upbeta}}_l + \widehat{\eta_l}(\mf{x}_u) = \mf{U}_l(\mf{x}_u)^\top \widehat{\bm{\upbeta}}_l + \sum_{i=1}^n (\widehat{\bm{\alpha}}_l)_i \kappa(\mf{x}_u, \mf{x}_i).
    \label{eq:mtlpredictor}
\end{equation}
Equation~\eqref{eq:mtlpredictor} shows that the prediction at any point is influenced by design points from all tasks. We note that the intrinsic noise matrix $\bm{\Sigma}_\eps$ appears explicitly in the SK predictor in Eq.~\eqref{eq:mseoptimalpredictor} but not in the MTL predictor in Eq.~\eqref{eq:mtlpredictor}. This is because the intrinsic noise matrix $\bm{\Sigma}_{\eps,l}$ for task $l$ affects the parameter estimate of $\bm{\alpha}_l$ implicitly, which is then used for prediction. These estimates are obtained using an EM algorithm which we derive in Section~\ref{sec:parameterestimation}.

%% file: 04parameterestimation.tex
\section{Parameter Estimation}\label{sec:parameterestimation}

The parameters of the complete model can be separated into three categories: (1) the intrinsic uncertainty $\mathsf{V}_l(\mf{x}_i)$ at each design point in each task, (2) the parameters $\{\bm{\upmu}_\alpha,\mf{C}_\alpha, \bm{\alpha}_l\}$ for the MTL formulation, and (3) the parameters $\bm{\upbeta}_l$ of the global trend for each task. We present an iterative method to simultaneously estimate the MTL parameters and the parameters of the global trend. We also briefly discuss a method to obtain point estimates for the hyperparameters $\{ \lambda, \nu, \bm{\uptheta} \}$, algorithm complexity, and stopping criteria. 

\subsection{Estimating the Intrinsic Uncertainty}\label{sec:intrinsicestimation}
In the single-task case, the intrinsic uncertainty at the design points was estimated using the sample variance at those points. We use a similar method for estimating the variance for many tasks by letting $\widehat{\mathsf{V}}_l(\mf{x}_i) = \widehat{\sigma}_{i,l}^2$, where
\begin{equation}
    \widehat{\sigma}_{i,l}^2 = \frac{1}{n_{i,l}-1}\sum_{j=1}^{n_{i,l}}\left( \mc{Z}_l^j(\mf{x}_i) - \overline{\mc{Z}}_l(\mf{x}_i) \right)^2.
\end{equation}
Hence, we can write the estimated intrinsic noise matrix for task $l$ as
\begin{equation}
    \widehat{\bm{\Sigma}}_{\eps,l} = \text{Diag}\{ \widehat{\sigma}_{1,l}^2 / n_{1,l}, \widehat{\sigma}_{2,l}^2 / n_{2,l}, \dots ,\widehat{\sigma}_{n_l,l}^2 / n_{n_l,l} \}.
    \label{eq:intrinsicnoisematrix}
\end{equation}

The above estimation method works in most real-world problems in manufacturing. For instance, While mapping key process input variables (KPIV) to key process output variables (KPOV), repeated runs of experiments are conducted at different design points; Eq.~\eqref{eq:intrinsicnoisematrix} can be used to estimate intrinsic uncertainty at each design point. Similarly, gauge R\&R studies are usually conducted before taking surface measurements; thus, the corresponding $\widehat{\sigma}^2$ are already available in the form of gauge repeatability (or precision). Equation~\eqref{eq:intrinsicnoisematrix} is generalizable to any number of experiments ($\geq$2) at any design point in any task, with arbitrarily large number of gauge resolutions. We use this $\widehat{\bm{\Sigma}}_{\eps,l}$ while estimating all other model parameters.

\subsection{Estimating the MTL Parameters Using EM}\label{sec:emestimation}
The parameters for the MTL model $\{ \bm{\upmu}_\alpha, \mf{C}_\alpha \}$ along with $\{ \widehat{\bm{\alpha}}_l : l = 1,2,\dots,m\}$ are learned using an EM algorithm. The complete EM algorithm is derived in Appendix. Here, we present the E and M steps briefly.

\textit{E-Step}:
Given the current $\bm{\upmu}_\alpha$ and $\mf{C}_\alpha$, estimate the expectation and covariance of $\bm{\alpha}_l , l = 1,2,...,m$.
\begin{equation}
    \widehat{\bm{\alpha}}_l = \left(\bm{\kappa}_l^\top \widehat{\bm{\Sigma}}_{\eps,l}^{-1} \bm{\kappa}_l + \mf{C}_\alpha^{-1} \right)^{-1} \left( \bm{\kappa}_l^\top \widehat{\bm{\Sigma}}_{\eps,l}^{-1} \bm{\upeta}_l + \mf{C}_\alpha^{-1}\bm{\upmu}_\alpha \right),
\end{equation}
\begin{equation}
    \mf{C}_{\alpha_l} = \left( \bm{\kappa}_l^\top \widehat{\bm{\Sigma}}_{\eps,l}^{-1} \bm{\kappa}_l + \mf{C}_\alpha^{-1} \right)^{-1},
\end{equation}
where $\bm{\kappa}_l \in \mb{R}^{n_l \times n}$ is the base kernel function ${\kappa} (\cdot , \cdot)$ evaluated between $\mf{X}_l$ and $\mf{X}$.

\textit{M-Step}:
Optimize $\bm{\upmu}_\alpha$ and $\mf{C}_\alpha$.
\begin{equation}
    \bm{\upmu}_\alpha = \frac{1}{\lambda + m} \sum_{l=1}^m \widehat{\bm{\alpha}}_l,
\end{equation}
\begin{align}
    \mf{C}_\alpha = \frac{1}{\nu + m} \left[ \vphantom{\sum_{l=1}^m} \lambda \bm{\upmu}_\alpha \bm{\upmu}_\alpha^\top + \nu \bm{\kappa}^{-1} + \sum_{l=1}^m \mf{C}_{\alpha_l} + \sum_{l=1}^m ( \widehat{\bm{\alpha}}_l - \bm{\upmu}_\alpha )(\widehat{\bm{\alpha}}_l - \bm{\upmu}_\alpha)^\top \right].
\end{align}

This EM algorithm is combined with the algorithm proposed for the regression parameters $\bm{\upbeta}_l$ in Section~\ref{sec:globaltrendestimation} to learn the complete model iteratively.

\subsection{Estimating the Parameters of the Global Trend}\label{sec:globaltrendestimation}
\begin{figure}[t]
    \centering
    \includegraphics[width=1\textwidth]{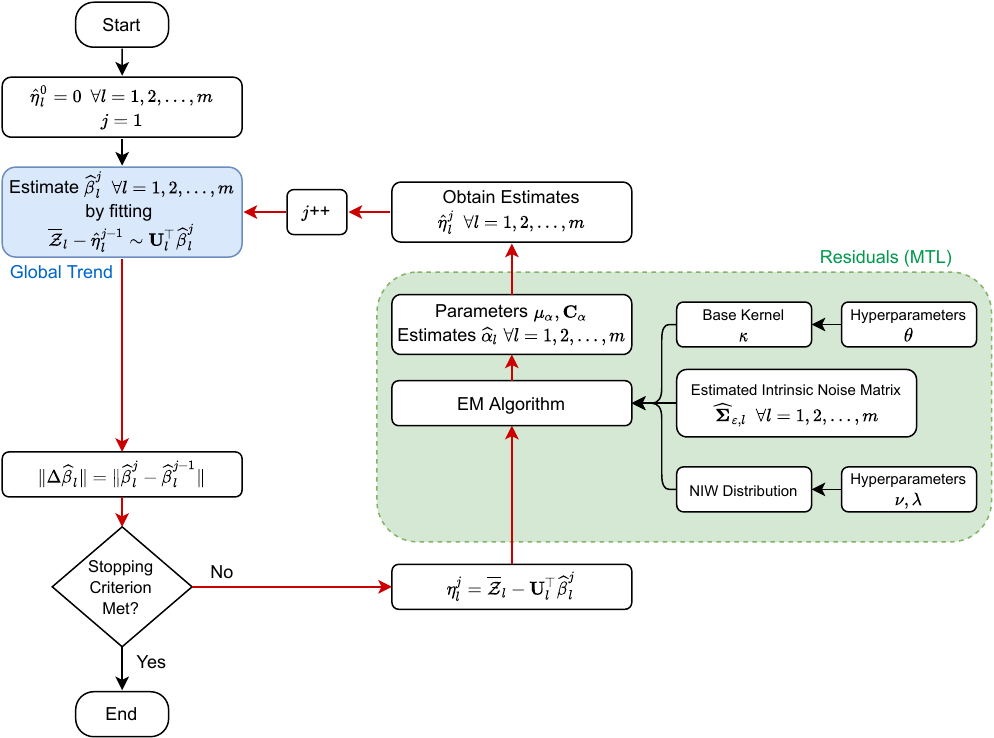}
    \caption{Flowchart for the iterative model update procedure for parameter estimation.}
    \label{fig:modelupdate}
\end{figure}

The parameters $\bm{\upbeta}_l$ for the global trend can be estimated using an iterative procedure which alternates between the global trend and the residuals. Such iterative models have been used for obtaining parameter estimates for universal kriging in single-task~\cite{Hengl2007RK, Kitanidis1993Covariance} and multi-task~\cite{Shao2017MTL, yang2021hybrid} settings. The $\bm{\upbeta}_l$ are estimated using robust regression, linear regression, or a task-specific model. We initialize by setting the local spatial variability to zero for all tasks. In the first iteration, the parameters of the global trend are estimated individually for each task and the residuals for all tasks are jointly learned using the EM algorithm presented above. In the $j\textsuperscript{th}$ iteration, the parameters $\widehat{\bm{\upbeta}}^j_l$ of the global trend are fitted after subtracting the estimated residuals from the previous iteration, i.e., $\overline{\mc{Z}}_l - \widehat{\bm{\upeta}}_l^{j-1} \sim \mf{U}_l^\top \widehat{\bm{\upbeta}}^j_l$. Subsequently, the MTL model $\widehat{\bm{\upeta}}_l^j$ is updated using the new residuals $\overline{\mc{Z}}_l - \mf{U}_l^\top \widehat{\bm{\upbeta}}^j_l$. Since we subtract the global trend before the EM algorithm in every iteration, the zero-mean assumption for the local spatial variability is not violated. This iterative model update is continued until a stopping criterion is satisfied. The complete procedure is shown in Figure~\ref{fig:modelupdate}.

\subsection{Estimating Hyperparameters}\label{sec:hyperparameterestimation}
An underlying assumption for the EM to work is that point estimates of the model hyperparameters $\{\nu, \lambda\}$ and kernel hyperparameters $\bm{\uptheta}$ are available. If not, we would treat these hyperparameters as sampled from their own parent distributions and a full-Bayesian treatment of the model would be required. However, we can find reliable point estimates for these hyperparameters through separate experiments on a subset of the training data. This technique has been implemented in previous MTL studies~\cite{Shao2017MTL, Haotian2021SpTmMTGP, yang2021hybrid} and is not re-iterated here for succinctness.

\subsection{Stopping Criteria and Algorithm Complexity}\label{sec:complexity}
The EM algorithm can be terminated when the change in $\bm{\upmu}_\alpha$ is below a predetermined threshold, i.e., $\lVert \Delta \bm{\upmu}_\alpha \rVert < t_1$. A more rigorous criterion can be to terminate when $\lVert \Delta \widehat{\bm{\alpha}}_l \rVert < t_2 \hspace{0.5em} \forall l=1,2,\dots,m$. Along similar lines, the iterative model update can be terminated when $\lVert \Delta \widehat{\bm{\upbeta}}_l \rVert < t_3 \hspace{0.5em} \forall l=1,2,\dots,m$, or when the average change in the parameters for all tasks is less than a predetermined threshold, i.e., $\lVert \Delta \widehat{\bm{\upbeta}} \rVert = \frac{1}{m}\sum\limits_{l=1}^m \lVert \Delta \widehat{\bm{\upbeta}}_l \rVert < t_4$. The numerical values for these thresholds can be determined based on the problem at hand.

The computational complexity of the EM algorithm is $\mc{O}(k_1mn^3)$, where $k_1$ is the number of iterations required for convergence, $m$ is the number of tasks, and $n$ is the number of distinct training points across all tasks (training sample size). The overall computational complexity of the iterative model update is $\mc{O}(k_1k_2mn^3)$, where $k_2$ is the number of iterations required to reach the stopping criterion for the iterative model update. For such iterative procedures, the extrinsic covariance learned from the regression residuals in a single iteration is often satisfactory, and is not different enough from the covariance learned after several iterations to affect the complete model~\cite{Kitanidis1993Covariance} . The iterative procedure can thus be run for a fixed small number of iterations. For all practical purposes, the computational complexity of the EM algorithm is dominant and the overall complexity can be approximated to $\mc{O}(k_1mn^3)$.

%% file: 05casestudies.tex
\section{Case Studies}\label{sec:casestudies}

In this section, we demonstrate the effectiveness of the proposed H-MT-MF framework through two case studies that represent different manifestations of multi-fidelity data. Specifically, the first case study provides a controlled synthetic setting with explicitly defined fidelity-dependent noise levels. The second case study utilizes real-world automotive engine surface measurements obtained using metrology systems with different spatial resolutions and repeatability characteristics (precisions). While such settings are often described as multi-resolution problems in surface metrology, they can be more generally interpreted as multi-fidelity data integration problems. In this work, measurement resolution and repeatability are treated as fidelity attributes, where higher-resolution gauges provide lower intrinsic variance at increased cost, and lower-resolution gauges provide noisier but more economical observations. Thus, multi-resolution metrology represents a practically important example.

\subsection{1D Illustrative Example}
Consider a simple 1D problem over 3 tasks, where the response variables $y_i$ for task $i$ are as follows
\begin{align*}
    \text{Task 1: } y_1 &= 0.1 + 0.1x + \sin{\frac{\pi x}{5}} + 0.2\sin{\frac{4\pi x}{5}}\\
    \text{Task 2: } y_2 &= 5.0 - 0.2x + \sin{\frac{\pi x}{5}} + 0.4\sin{\frac{4\pi x}{5}}\\
    \text{Task 3: } y_3 &= 0.3 + 0.3x + \sin{\frac{\pi x}{5}} + 0.3\sin{\frac{4\pi x}{5}}\\
\end{align*}
for $x \in [0, 20]$. This general functional form $y_i = a_i + b_ix + \sin{\pi x/5} + c_i\sin{4\pi x/5}$ is chosen such that each task has a specific global trend dictated by $(a_i, b_i)$, and a residual governed by the sinusoidal terms that can be learned jointly. Within each task, we take 10 measurements at random points in $[0, 20]$ with three replicates at each measurement. Each measurement is done using either (1) a less accurate, low resolution gauge with $\sigma_{\text{low-res}} = 0.2$, or (2) a more accurate, high resolution gauge with $\sigma_{\text{high-res}} = 0.05$. The response surfaces for all three tasks are learned using the H-MT-MF framework.

\begin{figure}[h]
    \centering
    \includegraphics[width=1\textwidth]{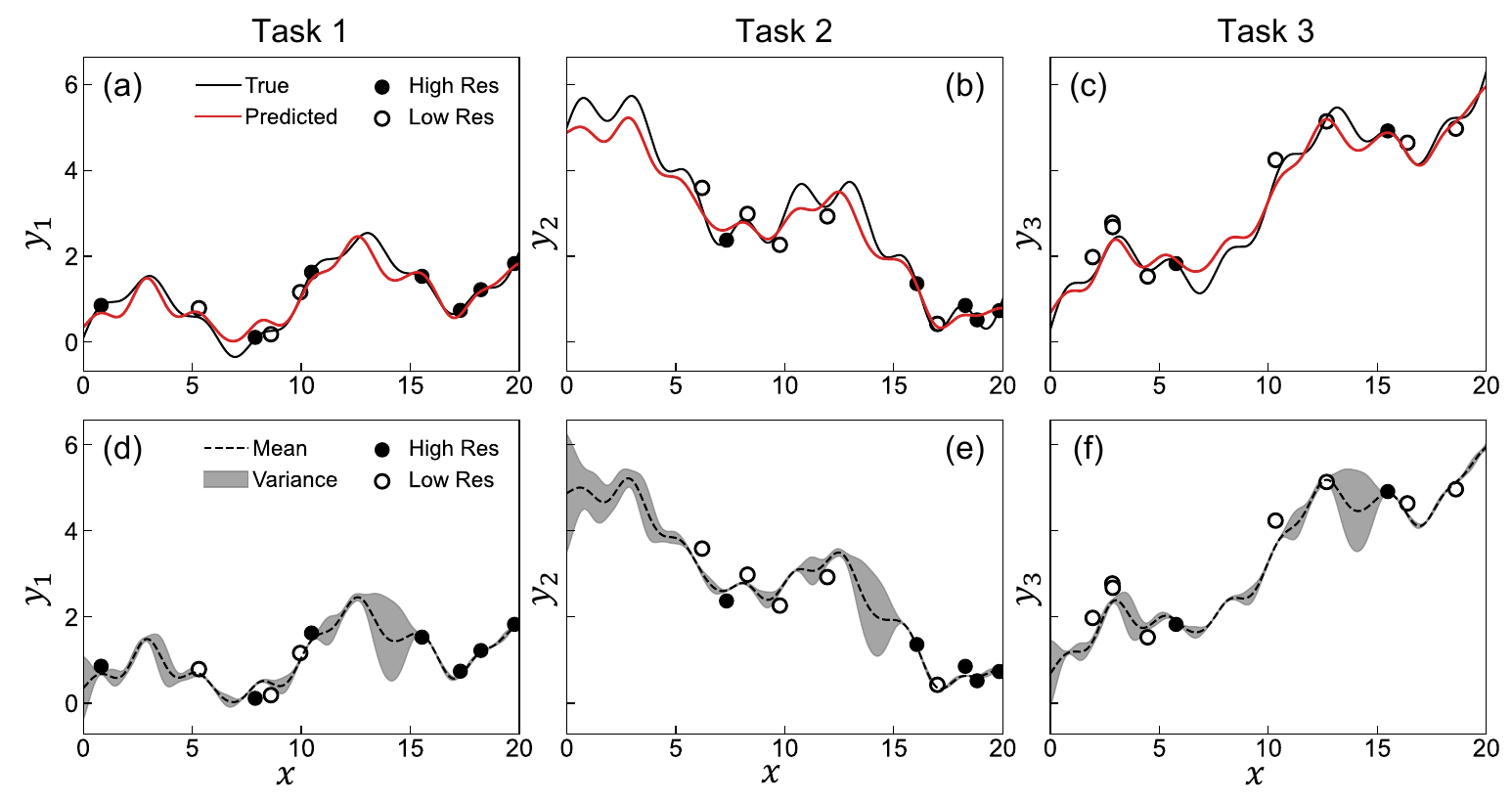}
    \caption{A 1D RSM example illustrating the H-MT-MF framework.}
    \label{fig:1DExample}
\end{figure}
Figures~\ref{fig:1DExample} (a), (b), and (c) show the true and predicted response surfaces for tasks 1, 2, and 3, respectively. Figures~\ref{fig:1DExample} (d), (e), and (f) show the posterior mean and variance for tasks 1, 2, and 3, respectively. Even though the underlying response surfaces follow complex, non-linear functions, the H-MT-MF framework closely predicts these functions in all three tasks with limited measurements. The low resolution measurements have larger deviations from the true function compared to the high resolution measurements; however, this intrinsic uncertainty does not adversely affect our predictions. The advantage of using MTL can be clearly seen in the transfer of information across tasks. For instance, there are no observations in the region $[0, 5]$ in task 2, but observations in tasks 1 and 3 indirectly influence task 2 predictions in the region resulting in better performance. The posterior variance (which measures prediction uncertainty) is also influenced across tasks -- observing regions of the domain in one task reduces the prediction uncertainty in the same region in all tasks. For instance, region and $[7, 10]$ in task 3 does not have any local measurements, but observing several points in tasks 1 and 2 leads to very low prediction uncertainty in this region in task 3. Finally, we see that the three tasks are similar-but-not-identical, i.e., they follow approximately similar residuals but vastly different global trends. The H-MT-MF framework is robust as it separates the global trend in each task and its residuals, thus allowing seemingly uncorrelated tasks to benefit from each other. 

\subsection{Case Study: Engine Surface Shape Prediction}
\label{sec:casestudysetup}
\begin{figure}[t]
    \centering
    \includegraphics[width=0.65\textwidth]{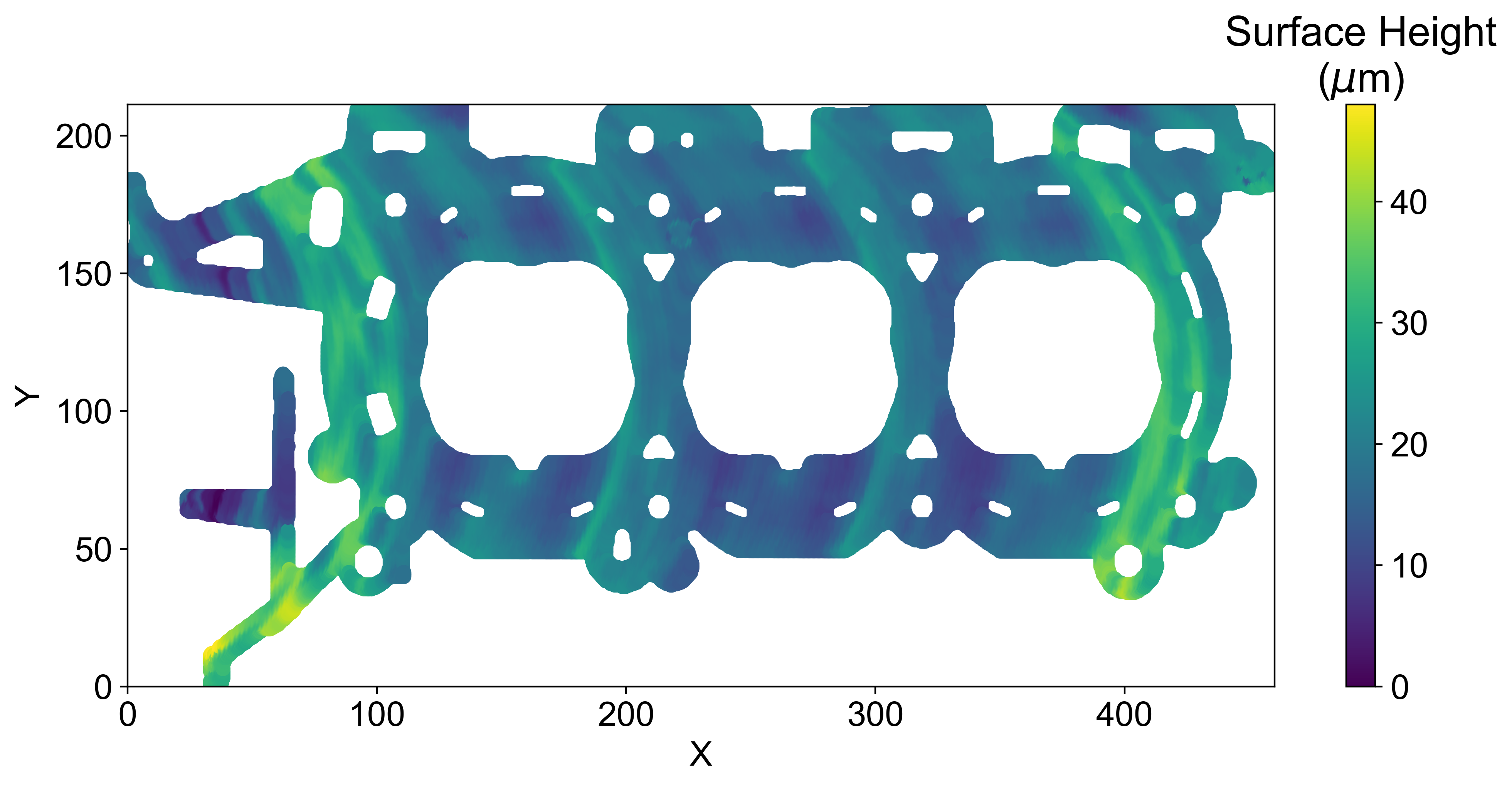}
    \caption{An example of an engine surface used in the case study.}
    \label{fig:surfaceexample}
\end{figure}
The data consists of three similar-but-not-identical engine block surfaces. Each engine surface is characterized by spatial coordinates $\mf{x}:=(x,y)$ and the corresponding surface height $Z(\mf{x})$. Figure~\ref{fig:surfaceexample} shows an engine surface example. The surface height at a location in the X-Y plane is depicted by the colorbar. The surfaces were machined using the same or parallel milling machines at a Ford engine plant~\cite{Shao2017MTL, Suriano2015}. As they come from the same parent manufacturing process, they satisfy the similar-but-not-identical property. The surface heights for all three surfaces are correlated and they also share a similar input space (as they are dimensionally similar). A task here is to learn the response surface for each engine surface from the set of points sampled from that surface.

We use a base kernel function $\kappa(\cdot, \cdot)$ of the squared exponential form as a measure of spatial similarity between two inputs. We choose this kernel function because of its popularity in spatial statistics and its demonstrated effectiveness in modeling surface shapes.
\begin{equation}
    \kappa(\mf{x}_i, \mf{x}_j) = \exp{\left(-\frac{\lVert \mf{x}_i - \mf{x}_j \rVert ^2 }{\delta^2}\right)}.
\label{basekernelcasestudy}
\end{equation}
The kernel hyperparameter $\delta^2$, along with the model hyperparameters $\nu$ and $\lambda$ are estimated using preliminary experiments on a subset of the training data, as explained in Section~\ref{sec:hyperparameterestimation}. For our case study, we use the point estimates $\delta^2 = 80$, $\nu = 1.0$, and $\lambda = 0.001$ for best model performance. Root Mean Squared Error (RMSE) is used as the metric for prediction accuracy. The RMSE for task $l$ is defined as
\begin{equation}
    \text{RMSE}_l = \sqrt{\frac{\sum\limits_{\mf{x} \in \mf{X}_{l,\text{test}}} \left( \widehat{\mathsf{Z}}_l (\mf{x}) - \mathsf{Z}_l (\mf{x}) \right)^2 }{ \lvert \mf{X}_{l,\text{test}} \rvert }},
\end{equation}
where $\mf{X}_{l,\text{test}}$ is the set of locations for task $l$ where predictions are made.


We measure 50 points (locations) randomly on each of the three engine surfaces and perform prediction on 15,000 points on each surface. From the 50 measured points, 25 are measured using a high resolution gauge with known repeatability $\sigma_\text{high-res}^2$, and 25 with a low resolution gauge with known repeatability $\sigma_\text{low-res}^2 > \sigma_\text{high-res}^2$. The observed value of the surface height $\mc{Z}(\mf{x})$ is a random sample drawn from a normal distribution with mean equal to the true surface height $Z(\mf{x})$ and variance equal to the gauge repeatability. Thus, the observed surface measurements differ from the true surface heights which better resembles a real measurement scenario. The intrinsic variance of the low and high resolution gauges is denoted as a percentage of the average surface height of the measured part. For instance, the average surface height of the part shown in Figure~\ref{fig:surfaceexample} is $\approx$20 $\mu$m. Thus, a gauge with $\sigma^2 = 0.1\%$ measures within 0.02 $\mu$m of the true surface height and is highly precise. Contrarily, a gauge with $\sigma^2 = 12.5\%$ only measures within 2.5 $\mu$m and is highly imprecise.

To formulate the basis functions $\mf{U}_l$ for all tasks, we use the Material Removal Rate (MRR) data which is cheaply available across the entire surface~\cite{Shao2017MTL}. The basis functions for task $l$ can be written as $\mf{U}_l(\mf{x}) = \begin{bmatrix} 1 & \text{MRR}_l(\mf{x}) \end{bmatrix}^\top$. This formulation stems from the fact that a higher MRR results in a larger displacement between the surface and the cutter, causing higher variations in surface height. For the case study, the MRR across the surface is simulated using the true surface height with a correlation of 0.7, i.e., $\rho(\text{MRR}_l, \mf{Z}_l) = 0.7 \hspace{0.5em}\forall l$. Since this relationship is linear, the parameters of the global trend $\bm{\upbeta}_l = \begin{bmatrix} \beta_0 & \beta_1 \end{bmatrix}_l^\top$ are fit using robust linear regression.

We compare the performance of the H-MT-MF framework to two other methods --- Engineering-Guided Multi-task Learning (EG-MTL)~\cite{Shao2017MTL} and SK~\cite{Ankenman2010StochasticKrig}. Like H-MT-MF, EG-MTL uses MTL to learn the local spatial variability across all surfaces together. However, the method assumes a homoscedastic intrinsic noise $\eps \sim \mc{N}(0,\sigma^2)$ across all design points and tasks. The parameter $\sigma^2$ is learned from the data; thus, the model is not capable of incorporating information about the gauge resolutions. On the other hand, SK incorporates gauge resolutions while making predictions; however, the response surface for each task is computed individually as there is no knowledge transfer between tasks. The parameter estimates for the SK model are obtained using maximum likelihood; we implement this using the $\textit{mlegp}$ package in $\textsc{R}$~\cite{Dancik2008Mlegp}. Table~\ref{tab:methodsummaryprediction} summarizes the comparison between the three methods.

\begin{table}[h]
    \centering
    \caption{Summary of the RSM methods used for comparison}
    \label{tab:methodsummaryprediction}
    \begin{tabular}{ccc}
    \toprule
        \makecell{Method\\} & \makecell{Knowledge Transfer Between\\Surfaces? (MTL)} & \makecell{Utilizes Fidelity Information?\\(Multi-fidelity)} \\
        \midrule
        H-MT-MF & Yes & Yes \\
        EG-MTL & Yes & No \\
        SK & No & Yes \\
        \bottomrule
    \end{tabular}
\end{table}

\begin{figure}[!t]
    \centering
    \includegraphics[width=1\textwidth]{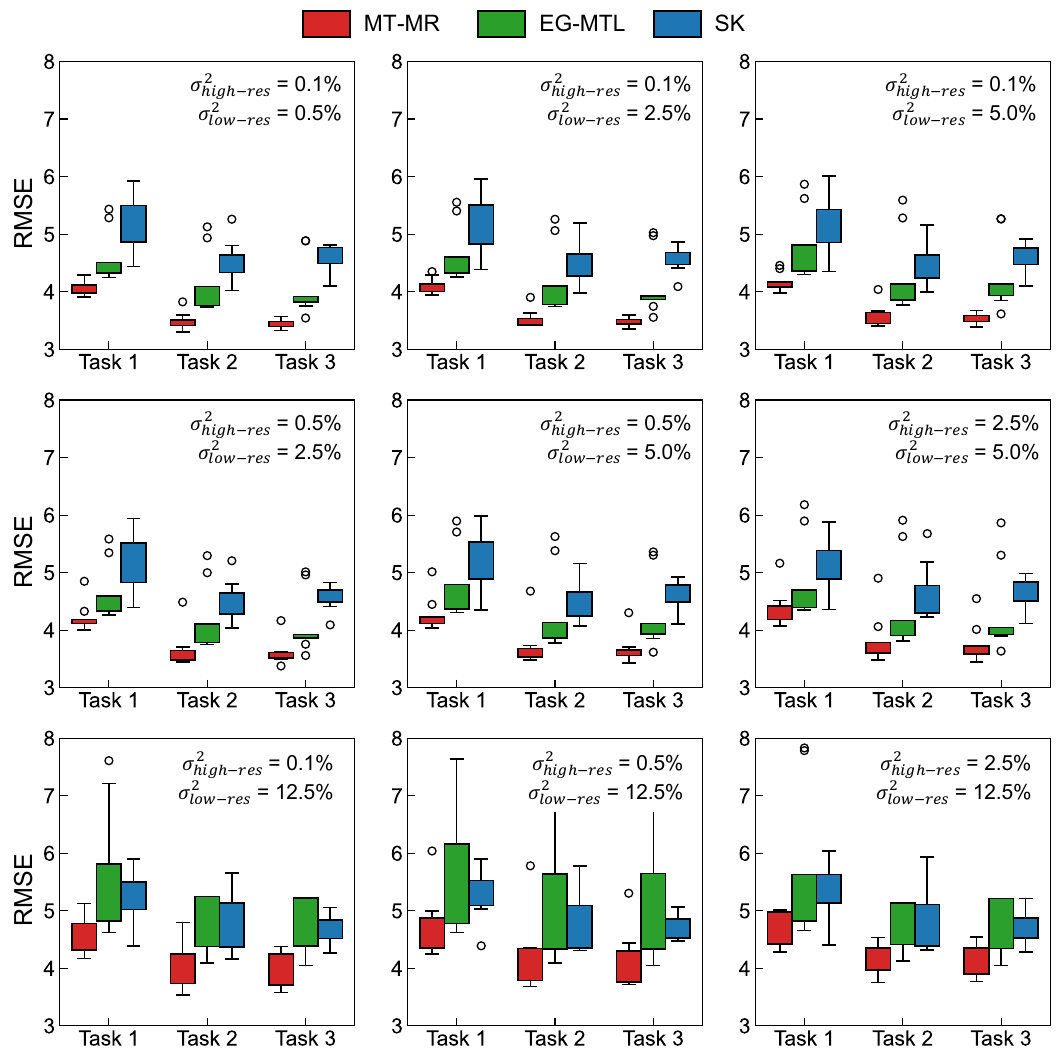}
    \caption{Results for the prediction comparison at different gauge resolutions. $\sigma_\text{high-res}^2$ and $\sigma_\text{low-res}^2$ are the repeatabilities of the high and low resolution gauges respectively, expressed as percentage of the average surface height.}
    \label{fig:predictionresults}
\end{figure}

To study the effect of the intrinsic uncertainty on prediction performance, we run each method for different pairs of low and high resolution gauges. We repeat each run ten times using different measurement locations randomly. Figure~\ref{fig:predictionresults} shows the RMSEs for nine different pairs of low and high-resolution gauges chosen in decreasing order of gauge precision. In the first pair ($\sigma_\text{high-res}^2 = 0.1\%$ and $\sigma_\text{low-res}^2 = 0.5\%$), both gauges are very precise. Contrarily, in the last pair ($\sigma_\text{high-res}^2 = 2.5\%$ and $\sigma_\text{low-res}^2 = 12.5\%$), both gauges are very imprecise. The rest of the gauge pairs are selected in between the above two such that a wide spectrum of intrinsic noise can be analyzed.

Figure~\ref{fig:predictionresults} shows that our H-MT-MF framework outperforms both EG-MTL and SK in all nine cases. This shows that incorporating gauge precision while learning tasks together yields the best prediction performance. The RMSEs for all three methods increase as gauge imprecision increases; however, the RMSEs for the H-MT-MF framework and SK are not severely affected as both methods account for intrinsic uncertainty in their model formulation. Contrarily, the RMSEs for EG-MTL increase rapidly with increase in intrinsic noise. For very high input uncertainties (bottom three plots), EG-MTL is even outperformed by SK which only learns tasks individually. Further, the robustness (indicated by the errorbar) of H-MT-MF and SK is not affected by gauge precision as much as EG-MTL. Lastly, the homoscedastic noise $\sigma^2$ learned from EG-MTL computes to $\approx$0.01\% of the surface mean, and is found to be the same for all nine cases analyzed. Physically, this means that EG-MTL assumes all points are sampled using an extremely precise gauge with negligible intrinsic noise, which is not the case. Our H-MT-MF framework thus resembles the real measurement system better while also learning tasks together.

\begin{table}[!t]
    \centering
    \caption{Average RMSE improvement for different gauge resolutions (fidelity levels)}
    \label{tab:improvementtable}
    \makebox[\textwidth][c]{%
    \begin{tabular}{cc|cccc|cccc}
    \toprule
    \multicolumn{2}{c|}{\thead{Gauge Resolutions}}
    & \multicolumn{4}{c|}{\thead{$\Delta_+\text{RMSE}_\text{EG-MTL}$}} & \multicolumn{4}{c}{\thead{$\Delta_+\text{RMSE}_\text{SK}$}}\\
    \midrule
    \thead{$\sigma_\text{high-res}^2$(\%)} & \thead{$\sigma_\text{low-res}^2$(\%)} & \thead{Task 1} & \thead{Task 2} & \thead{Task 3} & \thead{Average} & \thead{Task 1} & \thead{Task 2} & \thead{Task 3} & \thead{Average} \\
    \toprule
    0.1 & 0.5 & 10.76 & 14.33 & 14.17 & 13.09 & 21.14 & 22.21 & 24.50 & 22.62\\
    0.1 & 2.5 & 10.4 & 13.82 & 13.83 & 12.70 & 20.34 & 20.98 & 23.70 & 21.67\\
    0.1 & 5.0 & 11.35 & 14.69 & 14.81 & 13.62 & 19.02 & 19.67 & 22.95 & 20.55\\
    0.1 & 12.5 & 16.79 & 20.00 & 20.44 & 19.08 & 13.05 & 15.29 & 15.72 & 14.69\\
    \midrule
    0.5 & 2.5 & 8.33 & 11.23 & 11.12 & 10.23 & 18.41 & 18.29 & 21.08 & 19.26\\
    0.5 & 5.0 & 9.52 & 12.48 & 12.42 & 11.47 & 17.64 & 17.74 & 20.44 & 18.61\\
    0.5 & 12.5 & 14.34 & 17.09 & 17.28 & 16.24 & 11.96 & 12.67 & 12.70 & 12.44\\
    \midrule
    2.5 & 5.0 & 8.94 & 11.45 & 11.34 & 10.58 & 15.98 & 17.73 & 18.51 & 17.41\\
    2.5 & 12.5 & 15.00 & 18.44 & 17.59 & 17.01 & 12.74 & 14.14 & 12.51 & 13.13\\
    \bottomrule
    \end{tabular}
    }
\end{table}

We define another metric $\Delta_+ \text{RMSE}$ as the percentage improvement in RMSE compared to EG-MTL or SK. We define $\Delta_+ \text{RMSE}$ for EG-MTL as
\begin{equation}
    \Delta_+\text{RMSE}_\text{EG-MTL} = \frac{\text{RMSE}_\text{EG-MTL} - \text{RMSE}_\text{H-MT-MF}}{\text{RMSE}_\text{EG-MTL}}\times 100\%
\end{equation}
and similarly for SK. Table~\ref{tab:improvementtable} shows the average RMSE improvements in each task for the nine cases from Figure~\ref{fig:predictionresults}. $\Delta_+ \text{RMSE}_\text{EG-MTL}$ increases as the gauge precision decreases, which shows that including intrinsic uncertainty in the model formulation becomes more useful as the intrinsic noise increases. Contrarily, $\Delta_+ \text{RMSE}_\text{SK}$ decreases as the gauge precision decreases, for which we hypothesize two reasons -- (1) the similar-but-not-identical property of the tasks is affected which impacts learning performance, or (2) the observation errors across tasks accumulate which impacts prediction performance. Regardless of the trends, the performance of the H-MT-MF framework is consistently better than both EG-MTL and SK.

%% file: 06conclusion.tex
\section{Conclusion and Future Work}\label{sec:conclusion}

In this paper, we developed a novel H-MT-MF framework for data-efficient GP-based surrogate modeling in manufacturing. The proposed approach decomposes each response surface into two components: a task-specific global trend and a residual local variability component that is jointly learned across multiple related tasks. By simultaneously leveraging cross-task similarity and fidelity-dependent intrinsic uncertainty, the H-MT-MF framework enables principled integration of heterogeneous data sources. Furthermore, we provided a rigorous probabilistic derivation of the proposed model and developed a customized EM algorithm for efficient parameter estimation under the coupled cross-task and multi-fidelity structure. The H-MT-MF framework provides a unified and extensible statistical foundation for surrogate modeling in manufacturing systems characterized by multiple related processes and heterogeneous data sources.

Through both a synthetic 1D example and a real-world engine surface shape prediction case study, we demonstrated that jointly modeling tasks while incorporating fidelity-dependent uncertainty yields superior predictive accuracy and robustness. Compared with a state-of-the-art MTL model that assumes homogeneous noise and a SK model that treats tasks independently, the proposed framework consistently achieves improved performance across varying fidelity settings. 

Building up on this work, future research can focus on two areas. First, the proposed framework can be extended to spatiotemporal processes, which are ubiquitous in advanced manufacturing systems (e.g., ~\cite{Haotian2021SpTmMTGP, Wang2019Spatiotemporal, Wan2019Bayesian}). Recent studies have demonstrated the importance of spatiotemporal surrogate modeling for applications such as tool condition monitoring and production performance forecasting. Incorporating temporal dynamics into the H-MT-MF formulation would further enhance its applicability.

Second, intelligent sampling strategies can be developed to fully exploit the capabilities of the H-MT-MF framework while minimizing measurement cost and production disruption. In H-MT-MF settings, sampling design naturally becomes a hierarchical decision-making problem involving (1) selection of the task to sample, (2) selection of the spatial or process location within that task, and (3) selection of the fidelity level or measurement resolution. Developing adaptive, variance-based sampling strategies, which may be aided by heuristic search (e.g., hierarchical genetic algorithms \cite{Yang2019Hierarchical}) or Bayesian optimization techniques, would enable more efficient allocation of data acquisition resources.

%% file: appendix.tex
\appendix
\section{Derivation of the EM algorithm}\label{appendix:EM}
\noindent The hierarchical Bayes model presented for MTL can be written as
\begin{align*}
    \bm{\upmu}_\alpha, \mf{C}_\alpha &\sim \mc{N}\left(0,\frac{1}{\lambda}\mf{C}_\alpha \right)\mc{IW}(\nu, \bm{\kappa}^{-1}),\\
    \bm{\alpha}_l &\sim \mc{N}(\bm{\upmu}_\alpha, \mf{C}_\alpha),\\
    \eta_{i,l} &\sim \mc{N}\bigg(\bm{\kappa}(\mf{x}_i,\cdot)\bm{\alpha}_l \hspace{0.1em} , \hspace{0.2em} \widehat{\sigma}_{i,l}^2 \bigg),
\end{align*}
where $\eta_{i,l} = \eta_l(\mf{x}_i)$ for $\mf{x}_i \in \mf{X}_l$, and $\bm{\kappa}(\mf{x}_i, \cdot)$ is the $1\times n$ vector $\bm{\kappa}(\mf{x}_i, \mf{X})$. Further let $\upphi := \{\bm{\upmu}_\alpha, \mf{C}_\alpha\}$. The joint distribution of $\bm{\upeta}_l$ and $\bm{\alpha}_l$ given $\upphi$ can be written as
\begin{align*}
    p(\{\bm{\upeta}_l\}, \{\bm{\alpha}_l\} | \{\mf{X}_l\}, \upphi ) &= p(\{\bm{\upeta}_l\}| \{\bm{\alpha}_l\}, \{\mf{X}_l\}, \upphi)\cdot p( \{\bm{\alpha}_l\} | \upphi)\\
    &= \prod_{l=1}^m \frac{1}{\Lambda_l} \exp{\left(-\frac{1}{2} \Upsilon(\bm{\alpha}_l) \right)},
\end{align*}
where
\begin{align*}
    \Lambda_l &= (2\pi)^\frac{n_l+n}{2} \lvert\mf{C}_\alpha\rvert^\frac{1}{2} \prod\limits_{i\sim l}\widehat{\sigma_i},\\
    \Upsilon(\bm{\alpha}_l) &= \sum_{i\sim l} \frac{\left(\bm{\kappa}(\mf{x}_i,\cdot)\bm{\alpha}_l - \eta_{i,l} \right)^2}{\widehat{\sigma_i}^2} + (\bm{\alpha}_l - \bm{\upmu}_\alpha)^\top \mf{C}_\alpha^{-1}(\bm{\alpha}_l - \bm{\upmu}_\alpha).
\end{align*}
We can rewrite $\Upsilon(\bm{\alpha}_l)$ using the estimated noise matrix for each task $\widehat{\bm{\Sigma}}_{\eps,l}$ as
\begin{align*}
    \Upsilon(\bm{\alpha}_l) = \left(\bm{\kappa}_l\bm{\alpha}_l - \bm{\upeta}_l \right)^\top \widehat{\bm{\Sigma}}_{\eps,l}^{-1} \left(\bm{\kappa}_l\bm{\alpha}_l - \bm{\upeta}_l \right) + (\bm{\alpha}_l - \bm{\upmu}_\alpha)^\top \mf{C}_\alpha^{-1}(\bm{\alpha}_l - \bm{\upmu}_\alpha),
\end{align*}
where $\bm{\kappa}_l = \bm{\kappa}(\mf{X}_l,\mf{X}) \in \mb{R}^{n_l \times n}$. From the joint distribution, we see that the \emph{a posteriori} distribution of the latent $\bm{\alpha}_l$ is a product of $m$ Gaussian posteriori distributions. At the E-step, we can compute the sufficient statistics of each Gaussian. The expectation and covariance of $\bm{\alpha}_l$ are obtained as shown.
\begin{align*}
    \frac{\partial \Upsilon(\bm{\alpha}_l)}{\partial\bm{\alpha}_l}=0 \implies \widehat{\bm{\alpha}}_l = \left(\bm{\kappa}_l^\top \widehat{\bm{\Sigma}}_{\eps,l}^{-1} \bm{\kappa}_l + \mf{C}_\alpha^{-1} \right)^{-1} \left( \bm{\kappa}_l^\top \widehat{\bm{\Sigma}}_{\eps,l}^{-1} \bm{\upeta}_l + \mf{C}_\alpha^{-1}\bm{\upmu}_\alpha \right),
\end{align*}
\begin{align*}
    \mf{C}_{\alpha_l} = \left(\frac{\partial^2 \Upsilon(\bm{\alpha}_l)}{\partial\bm{\alpha}_l\partial\bm{\alpha}_l^\top} \right)^{-1} = \left(\bm{\kappa}_l^\top \widehat{\bm{\Sigma}}_{\eps,l}^{-1} \bm{\kappa}_l + \mf{C}_\alpha^{-1} \right)^{-1}.
\end{align*}
The log-likelihood of the complete data is
\begin{multline*}
    \ln p(\{\bm{\upeta}_l\}, \{\bm{\alpha}_l\} | \{\mf{X}_l\}, \upphi ) = -\frac{1}{2}\sum_{l=1}^m \bigg[(n_l+n)\ln(2\pi) + \sum\limits_{i\sim l}\ln(\widehat{\sigma_i}^2) +\ln\lvert\mf{C}_\alpha\rvert \\ + \left(\bm{\kappa}_l\bm{\alpha}_l - \bm{\upeta}_l \right)^\top \widehat{\bm{\Sigma}}_{\eps,l}^{-1} \left(\bm{\kappa}_l\bm{\alpha}_l - \bm{\upeta}_l \right) + (\bm{\alpha}_l - \bm{\upmu}_\alpha)^\top \mf{C}_\alpha^{-1}(\bm{\alpha}_l - \bm{\upmu}_\alpha)\bigg].
\end{multline*}
Define $Q(\upphi)$ as the expected value of the log likelihood w.r.t. the current conditional distribution of $\bm{\alpha}_l$, i.e., $ Q(\upphi) = \mb{E}_{\bm{\alpha}_l|\bm{\upeta}_l, \{\mf{X}_l\}, \upphi} [\ln p(\{\bm{\upeta}_l\}, \{\bm{\alpha}_l\} | \{\mf{X}_l\}, \upphi )]$. Then
\begin{multline*}
    Q(\upphi) = -\frac{1}{2}\left[ \sum_{l=1}^m(n_l+n)\ln(2\pi) + \sum\limits_{i=1}^n\ln(\widehat{\sigma_i}^2)\right] - \frac{m}{2}\bigg[ \ln\lvert\mf{C}_\alpha\rvert + \bm{\upmu}_\alpha^\top \mf{C}_\alpha^{-1} \bm{\upmu}_\alpha  \bigg] \\ -\frac{1}{2}\sum_{l=1}^m \bm{\upeta}_l^\top \widehat{\bm{\Sigma}}_{\eps,l}^{-1} \bm{\upeta}_l -\frac{1}{2}\sum_{l=1}^m 
    \bigg\{ 
    \mb{E}\bigg[ \bm{\alpha}_l^\top \left(\bm{\kappa}_l^\top \widehat{\bm{\Sigma}}_{\eps,l}^{-1} \bm{\kappa}_l + \mf{C}_\alpha^{-1} \right) \bm{\alpha}_l \bigg]\\
    - 2\mb{E}\bigg[ \bm{\alpha}_l^\top \left( \bm{\kappa}_l^\top \widehat{\bm{\Sigma}}_{\eps,l}^{-1} \bm{\upeta}_l + \mf{C}_\alpha^{-1}\bm{\upmu}_\alpha \right) \bigg] 
    \bigg\}.
\end{multline*}
We evaluate the two expectations in the above equation using the following results: let $\mf{x}$ be an $n\times 1$ random vector with $\mb{E}[\mf{x}] = \upmu$ and $\text{Var}[\mf{x}] = \Sigma$, $\mf{A}$ be an $n \times n$ symmetric matrix and $\mf{B}$ be an $n \times n$ constant matrix. Then, $\mb{E}[\mf{x}^\top \mf{A} \mf{x}] = \text{tr}(\mf{A}\Sigma) + \upmu^\top \mf{A} \upmu$, and $\mb{E}[\mf{x}^\top \mf{B}] = \upmu^\top \mf{B}$. Thus,
\begin{multline*}
    Q(\upphi) = -\frac{1}{2}\left[ \sum_{l=1}^m(n_l+n)\ln(2\pi) + \sum\limits_{i=1}^n\ln(\widehat{\sigma_i}^2)\right] - \frac{m}{2}\bigg[ \ln\lvert\mf{C}_\alpha\rvert + \bm{\upmu}_\alpha^\top \mf{C}_\alpha^{-1} \bm{\upmu}_\alpha  \bigg] \\ -\frac{1}{2}\sum_{l=1}^m \bm{\upeta}_l^\top \widehat{\bm{\Sigma}}_{\eps,l}^{-1} \bm{\upeta}_l -\frac{1}{2}\sum_{l=1}^m 
    \bigg\{ \text{tr}\left( \big(\bm{\kappa}_l^\top \widehat{\bm{\Sigma}}_{\eps,l}^{-1} \bm{\kappa}_l + \mf{C}_\alpha^{-1} \big) \mf{C}_{\alpha_l}  \right) \\+ \widehat{\bm{\alpha}}_l^\top \left(\bm{\kappa}_l^\top \widehat{\bm{\Sigma}}_{\eps,l}^{-1} \bm{\kappa}_l + \mf{C}_\alpha^{-1} \right) \widehat{\bm{\alpha}}_l
    -2\widehat{\bm{\alpha}}_l^\top \left( \bm{\kappa}_l^\top \widehat{\bm{\Sigma}}_{\eps,l}^{-1} \bm{\upeta}_l + \mf{C}_\alpha^{-1}\bm{\upmu}_\alpha \right) 
    \bigg\}.
\end{multline*}
We can rewrite $Q(\upphi)$ as
\begin{multline*}
    Q(\upphi) = \text{const} - \frac{m}{2}\bigg[ \ln\lvert\mf{C}_\alpha\rvert + \bm{\upmu}_\alpha^\top \mf{C}_\alpha^{-1} \bm{\upmu}_\alpha  \bigg] \\ -\frac{1}{2}\sum_{l=1}^m 
    \bigg\{ \text{tr}\left( \mf{C}_\alpha^{-1} \mf{C}_{\alpha_l}  \right)
    + \widehat{\bm{\alpha}}_l^\top \mf{C}_\alpha^{-1} \widehat{\bm{\alpha}}_l - 2\widehat{\bm{\alpha}}_l^\top \mf{C}_\alpha^{-1} \bm{\upmu}_\alpha \bigg\},
\end{multline*}
where `$\text{const}$' encompasses all the terms that do not depend on $\upphi$. In the M-step, the updates for $\upphi$ can be obtained by maximizing the penalized likelihood of the complete data, i.e., $\arg\max_\upphi p(\{\bm{\upeta}_l\}, \{\bm{\alpha}_l\} | \{\mf{X}_l\}, \upphi ) \cdot p(\upphi) = \arg\max_\upphi Q(\upphi) + \ln p (\upphi)$, where $p(\upphi) = p(\bm{\upmu}_\alpha, \mf{C}_\alpha)$ is the hyperprior with density
\begin{align*}
    p(\bm{\upmu}_\alpha, \mf{C}_\alpha) \propto \lvert\mf{C}_\alpha\rvert^{\frac{\nu+n}{2}+1} \exp\left( -\frac{1}{2}\text{tr}(\bm{\kappa}^{-1}\mf{C}_\alpha^{-1}) -\frac{\lambda}{2}\bm{\upmu}_\alpha^\top \mf{C}_\alpha^{-1} \bm{\upmu}_\alpha \right).
\end{align*}
Solving the above maximization is straightforward and yields the following updates for $\bm{\upmu}_\alpha$ and $\mf{C}_\alpha$:
\begin{align*}
    \bm{\upmu}_\alpha = \frac{1}{\lambda + m} \sum_{l=1}^m \widehat{\bm{\alpha}}_l,
\end{align*}
\begin{align*}
    \mf{C}_\alpha = \frac{1}{\nu + m} \left[ \vphantom{\sum_{l=1}^m} \lambda \bm{\upmu}_\alpha \bm{\upmu}_\alpha^\top + \nu \bm{\kappa}^{-1} + \sum_{l=1}^m \mf{C}_{\alpha_l} + \sum_{l=1}^m ( \widehat{\bm{\alpha}}_l - \bm{\upmu}_\alpha )(\widehat{\bm{\alpha}}_l - \bm{\upmu}_\alpha)^\top \right].
\end{align*}